\documentclass{article}
\usepackage[utf8]{inputenc}

\usepackage{float}
\usepackage{graphicx}
\usepackage{amsmath}
\usepackage{amssymb}
\usepackage{color}
\usepackage{xcolor}

\usepackage{calc}

\DeclareMathOperator*{\argmin}{\arg\!\min\;}

\title{Towards a Better Understanding of Predict and Count Models}
\author{
S. Sathiya Keerthi\thanks{Cloud \& Information Services Lab, Microsoft, Mountain View, CA 94043}
\and
Tobias Schnabel \thanks{Department of Computer Science, Cornell University, Ithaca, NY 14853}
\and
Rajiv  Khanna \thanks{Department of Electrical and Computer Engineering, The University of Texas at Austin, Austin, TX 78712}
}

\date{\today}

\begin{document}

\newtheorem{observation}{Observation}

\def\mp{\mbox{\#}}
\def\ltil{\tilde{\ell}}
\def\Ctil{\tilde{C}}
\def\w{{\mathbf w}}
\def\c{{\mathbf c}}

\def\W{{\mathbf W}}
\def\C{{\mathbf C}}
\def\Cc{{\cal C}}
\def\N{{\cal N}}

\def\e{{\bf e}}
\def\spmi{{\textrm{\tiny SPMI}}}
\def\sppmi{{\textrm{\tiny SPPMI}}}
\def\pmi{{\textrm{\tiny PMI}}}
\def\Pmi{{\textrm{PMI}}}
\def\V{{\cal V}}
\def\svd{{\textrm{\tiny SVD}}}
\def\ssvd{{\textrm{\tiny SSVD}}}
\def\full{{\textrm{\tiny Full}}}
\def\symm{{\textrm{\tiny Symm}}}
\def\z{{\bf z}}
\def\neg{{\textrm{\tiny neg}}}

\def\x{{\bf x}}

\newcommand{\T}[1]{\par\textcolor{violet}{\textbf{Tobias:} #1}}
\newcommand{\K}[1]{\par\textcolor{violet}{\textbf{Keerthi:} #1}}

\newcommand*{\MyDef}{\mathrm{def}}
\newcommand*{\eqdefU}{\ensuremath{\mathop{\overset{\MyDef}{=}}}}
\newcommand*{\eqdef}{\mathop{\overset{\MyDef}{\resizebox{\widthof{\eqdefU}}{\heightof{=}}{=}}}}

\maketitle

\begin{abstract}
In a recent paper, Levy and Goldberg~\cite{Levy2014} pointed out an interesting connection between prediction-based word embedding models and count models based on pointwise mutual information. Under certain conditions, they showed that both models end up optimizing equivalent objective functions. This paper explores this connection in more detail and lays out the factors leading to differences between these models. We find that the most relevant differences from an optimization perspective are (i) predict models work in a low dimensional space where embedding vectors can interact heavily; (ii) since predict models have fewer parameters, they are less prone to overfitting.

Motivated by the insight of our analysis, we show how count models can be regularized in a principled manner and provide closed-form solutions for L1 and L2 regularization. Finally, we propose a new embedding model with a convex objective and the additional benefit of being intelligible. 
\end{abstract}

\section{Introduction}
\label{sec:intro}

Distributional semantic models~\cite{Turney2010}, also called \emph{count models}~\cite{Baroni2014} have been popular in the computational linguistics literature for several decades. In the last few years, however, \emph{predict models} such as the Skip-Gram model and Continuous Bag-Of-Words model have become the de facto standard for word modeling~\cite{Mikolov2013a, Mikolov2013b}. These methods have origins in neural language modeling~\cite{Mnih2009}, where the goal was to improve classic language models. A recent evaluation study~\cite{Baroni2014} suggested that the predict models are superior to the count models on a range of word similarity tasks. Newer work, however, attributes a large amount of the differences in performance between count and predict models to a lack of proper hyperparameter optimization~\cite{Levy2015}. This has prompted the need for understanding the differences between the two types of models.

To this end, Levy and Goldberg~\cite{Levy2014} made an interesting connection between two key models: a traditional count model based on pointwise mutual information (PMI) and a predict model, namely the skip-gram model with negative sampling (SGNS). The key result is that the SGNS model is equivalent to a shifted version of the PMI method, where all values get shifted by a factor of $\log k$. However, the proof assumes that the input and output dimensions are the same; hence the word vectors will still be of the size of the vocabulary. As we show in our analysis, this assumption has important implications for subsequent steps, such as dimensionality reduction methods like singular value decomposition (SVD). 


The aim of this paper is to analyze connections between count and predict models in a more detailed fashion. In particular, we investigate the differences between SGNS and PMI methods more deeply. We make several observations that help this understanding. The first useful result is that the Shifted PMI model comes out as an explicit version of SGNS in which the context word vectors are fixed at their one-hot representations. We point out two differences between the Shifted PMI model and the SGNS model from an optimization perspective. First, SGNS usually works in lower dimensions, making it possible for all word vectors two interact. Second, as count models have usually more parameters, this might cause them to overfit on the data. 

We use the insight from our analysis to propose several interesting extensions to classic word embedding models. For example, our analysis allows regularization to be added into PMI methods in a systematic way. We also propose a new convex word vector model based on CBOW which offers interpretability and a well-defined training objective. 

In short, we draw connections between existing count and prediction models and augment them in several ways. However, in order to fully understand the factors that differentiate various methods, a comprehensive set of experiments is needed. We plan to carry out these experiments as part of future work.

\section{Notations}
\label{sec:notns}
Consider a corpus $D = \{w^t\}_{t=1}^{|D|}$ that is a sequence of words over a vocabulary $\V$. We will use $w$ when referring to the index of a target word and $c$ when referring to the index of a context word. We define the exact notion of context in  Section~\ref{sec:context}. Note that $w$ and $c$ are integers taking values from $1$ to $|\V|$. 

Each word and context word has a vectorial representation. We will use $\w \in  \mathbb{R}^m$ and $\c \in  \mathbb{R}^n$ to denote the word vector and context vector corresponding to word $w$ and context word $c$. For vector ${\mathbf x}$, ${\mathbf x}_i$ will denote the $i$-th component of ${\mathbf x}$.  Let $\{\w\}$ and $\{\c\}$ denote the set of all word and context word vectors.
One can also think of $\{\w\}$ and $\{\c\}$ as matrices $\W$ and $\C$ whose $w$-th and $c$-th rows are given by $\w$ and $\c$ respectively. We will use $\W$ and $\C$ to synonymously refer to $\{\w\}$ and $\{\c\}$.

Let us use \# to denote a count and weighting function. In the simplest case, for example, \#$(w,c)$ could denote the number of occurrences of the (word, context word) pair $(w,c)$ in corpus $D$. 
Let us define $\mp (w,\cdot) = \sum_{c\in \V} \mp(w,c)$ and $\mp (\cdot,c) = \sum_{w\in \V} \mp(w,c)$, the total counts for word $w$ and context vector $c$. 


\subsection{Defining context}
\label{sec:context}
An important design choice in count models is the definition of \emph{context}. The definition of context determines which co-occurrences get considered when creating a model, along with the importance of single co-occurrences. A common way of defining context is based on choosing a window around each occurrence of a word $w$ in $D$. There are many options to customize this window; for example, it can either be symmetric or asymmetric around $w$. There could also exist weights (e.g., dependent on the relative positions of $w$ and $c$) associated with each $(w,c)$ occurrence in $D$. One may also include down-weighting of common words, possibly for both for the current word $w$ as well as for the context words $c$\footnote{Mikolov et al~\cite{Mikolov2013b} used down-weighting of common words. It is unclear if Levy and Goldberg~\cite{Levy2014} used it.}. All these choices determine the value of \#$(w,c)$. Note that since the weightings can be real, \#$(w,c)$, $\mp (w,\cdot)$ and $\mp (\cdot,c)$ can take non-negative real values in general. 

If the window is symmetric around $w$ and as well as the weighting functions are symmetric, then
\begin{enumerate}
\item $\mp(w,c)$ is symmetric, i.e., $\mp(w,c)=\mp(c,w)$ and
\item $\mp (w,\cdot) = \mp (\cdot,c)$ if $w$ and $c$ denote the same word
\end{enumerate}

We will refer to this as the {\it symmetry condition}. 
In Section~\ref{subsec:symm} we will revisit this condition and see that it helps our understanding of non-uniqueness and interchangeability of word vectors and context word vectors.

{\bf Example.} Consider an asymmetric window $K$ that includes $l$ words to the left and $r$ words to the right of a word $w$. In other words,
$$K=(-l,\cdots, -1, 1, \ldots, r).$$
Take one (word, context word) pair $(w^t,w^{t+i})$, where $i\in K$. Let $P(w^t)$ be the probability with which word $w^t$ is chosen for inclusion into the training set. For example, Mikolov et al~\cite{Mikolov2013b} suggest the following down-sampling probability:
\begin{equation}
\label{downsamp}
P_1(w^t) = 1 - \sqrt{\frac{\tau}{f(w^t)}}
\end{equation}
where $f(w^t)$ is the frequency of word $w^t$ in the corpus $D$, and $\tau$ is a threshold. Let $P_2(w^{t+i})$ be the weight associated with $w^{t+i}$ as context word. Here again we may want to down-sample frequent words similar to~(\ref{downsamp}). Let $P_3(i)$ be the weight associated with the position of the context word $w^{t+i}$ in the context window. Using all these individual weights we can define the overall weight for the $(w^t,w^{t+i})$ pair:
\begin{equation}
P(w^t,w^{t+i}) = P_1(w^t) P_2(w^{t+i}) P_3(i)
\end{equation}
Accumulating all this info over the entire corpus gives us the total value for (word, context word) pairs:
\begin{equation}
\label{countver}
\mp(w,c) = \sum_{\{(t,i): w^t=w, w^{t+i}=c\} } P(w^t, w^{t+i})
\end{equation}
Now \#$(w,\cdot)$ and \#$(\cdot,c)$ can be formed using \#$(w,c)$ as described in the beginning of this section.

\section{Background}
Remember that the goal of this paper is to investigate the differences between count and prediction models. To do this, we start by summarizing two special instances of count and prediction models. We first introduce PMI models as representatives of count models in Section~\ref{sec:pmi}, and then discuss the Skip-Gram model with Negative Sampling (SGNS) in Section~\ref{sec:sgns}.

\subsection{PMI models}
\label{sec:pmi}
PMI models have been used extensively in distributional semantic models~\cite{Turney2010} to compute similarities between words. As the name implies, the key quantity in these models is {\em Pointwise Mutual Information (PMI)}. Its definition and MLE estimate from data are given by
\begin{equation}
\Pmi(w,c) = \log \frac{P(w,c)}{P(w)P(c)} \approx \log \frac{\mp(w,c)\cdot |D|}{\mp(w,\cdot)\cdot\mp(\cdot,c)},
\label{pmidef}
\end{equation}
where $\mp(w,c)$ is the simple count function.
For a given word $w$, PMI represents $w$ by forming a vector whose components are $\Pmi(w, c)$ for all $c \in V$.

Note that PMI is not defined when $\mp(w,c)=0$. To circumvent this problem, Positive PMI (PPMI) replaces all negative values, i.e., $$\textrm{PPMI}(w, c) = \max(\Pmi(w,c),0).$$

More recently, Levy and Goldberg~\cite{Levy2014} have defined shifted variants of PMI and the PPMI metrics. Shifted PMI (SPMI) is just defined as $\Pmi(w,c) -\log k$, and Shifted Positive PMI (SPPMI) is defined as $\max(\text{SPMI}(w,c),0)$ where $k$ is some positive integer. PMI and PPMI are, respectively, special cases of Shifted PMI and Shifted PPMI, with $k$ set to $1$. We will return to Shifted PMI and Shifted PPMI in the next section, where we talk about the equivalence of certain methods.

As all vectors produced by PMI have the vocabulary size as the dimensionality, dimensionality reduction techniques are often applied to the original matrix to decrease the memory and computational requirements. 

\subsection{Skip-Gram with Negative Sampling (SGNS)}

SGNS~\cite{Mikolov2013b} is a popular predict model that aims to predict which word $w$ occurs with a context word $c$. One can derive the SGNS~\cite{Mikolov2013b} model from a binary classification setting where the target variable $y$ specifies whether a word $w$ occurs with context word $c$. SGNS tries to solve this task with the logistic loss applied to the input score $x=\w\cdot\c$. 

Remember that our corpus $D$ is a sequence of words, $D = \{w^t\}_{t=1}^{|D|}$. To make notation easier, here we assume that the context is defined by all the words within a window ${C^t}$ around each word $w^t$. Let the corresponding sequence of sets of context words be $\{C^t\}_{t=1}^T$. To construct the training set for the SGNS model, one forms one training example for each $t$ and each context word $c^t \in C^t$. For each $c^t\in C^t$, a set of $k$ negative context words, $\N_{c^t}^t$ is chosen at random and, the logistic loss function $f(\w_t,\c_t;\N_{c^t}^t)$ is applied. This loss is the aggregation of losses for one positive example and $k$ negative examples. More formally, the loss of one training example $\ell(\w^t,C^t)$ is:
\begin{align}
& \ell (\w^t,C^t) = \sum_{c^t\in C^t} f(\w^t,\c_t;\N_{c^t}^t) \eqdefU L(\w^t\cdot\c^t,1) + \sum_{c\in \N_{c^t}^t} L(\w^t\cdot\c,-1),
\end{align}
where $L(\w\cdot\c,y)$ is the logistic loss corresponding to the word-context pair $(w,c)$ and target $y\in\{1,-1\}$.
The loss over the entire corpus is then the sum of all individual losses:
\begin{align}
& \ell (\W, \C) = \sum_{t=1}^{|D|} \ell (\w^t,C^t).
\label{corpusform}
\end{align}


Let us refer to this way of writing the training objective (sum over occurrences in the corpus) as the {\it corpus format}.

Levy and Goldberg~\cite{Levy2014} showed that, by accumulating data over co-occurrences of words and context words, the objective function in (\ref{corpusform}) can be rewritten as\footnote{Levy and Goldberg~\cite{Levy2014} write the problem as the maximization of likelihood; hence the $\ell$ here and the one in~\cite{Levy2014} are negatives of each other.}:
\begin{eqnarray}
\ell (\W,\C) & = &  \sum_{w\in V} \sum_{c\in V} \ell_{w,c}(\w,\c) \label{ellall} \\
\ell_{w,c}(\w,\c) & = & \mp(w,c) L (\w\cdot \c, 1) + k\cdot \frac{\mp(w,\cdot)\cdot\mp(\cdot,c)}{|D|} L (\w\cdot \c,-1) \label{ellwc} 
\end{eqnarray}
Let us refer to this (equivalent) way of writing the training objective as the {\it co-occurrence format}. Note that in the co-occurrence format, we do not compute losses for each token individually, but instead compute a loss for each unique $(w,c)$ pair.

In \eqref{ellwc} we will from now on also consider other loss functions $L(x,y)$ apart from the logistic loss like in the original formulation. We consider general loss functions, $L(x,y)$ (here $y\in\{1,-1\}$ is the target variable) with the logistic loss being a special case. This generalization allows us to define custom loss functions that can incorporate domain knowledge or other side information. Table~\ref{tab:losslist} lists a set of popular loss functions.

\label{sec:sgns}
\begin{table}[bt]
\begin{center}
\begin{tabular}{|c|c|}
\hline
Loss name & $L(x,y)$ \\
\hline
\hline
Logistic & $\log (1 + e^{-yx}) $  \\
\hline
Squared & $\frac{1}{2}(x-y)^2$ \\
\hline
Squared Hinge & $L=\frac{1}{2}(\max(1-yx,0))^2$ \\
\hline
Hinge & $\max(1-yx,0)$  \\
\hline
Huber & $\frac{1}{2}(\max(1-yx,0))^2$ if $yx\ge -1$; $-2yx$ otherwise  \\
\hline
\end{tabular}
\caption{A list of popular loss functions, $L$.}
\label{tab:losslist}
\end{center}
\end{table}


\begin{observation}
Unlike SGNS, not all models can be reduced to the co-occurrence format - an example is the CBOW model~\cite{Mikolov2013a}. In fact, even with the skip-gram model, the reduction to the co-occurrence format is not possible if we use the traditional softmax or hierarchical softmax~\cite{Mnih2009} approach to speed up training. In general, models reducible to the co-occurrence format have to only be based on counts of words, context words and their co-occurrences. That property makes them have close relations with {\it count models}~\cite{Baroni2014,Levy2014,Shi2014}. Another model that can be expressed in co-occurrence form is Glove~\cite{glove}.
\end{observation}

\subsubsection{Remarks on symmetry}
\label{subsec:symm}

In case we assume symmetric windows as outlined in Section~\ref{sec:context} for counting co-occurrences,   we can swap word and context vectors at optimality. 

\begin{observation}
\label{obs:swap}
Suppose $(\W^*$, $\C^*)$ is a minimizer of $\ell$. Then $(\C^*$, $\W^*)$ is also a minimizer of $\ell$; in other words, at optimality, if the word vectors and context word vectors are completely swapped, optimality still holds. Thus, depending on how the numerical optimization of~(\ref{ellall}) initialized, $\W^*$ and $\C^*$ can end up being completely swapped.
\end{observation}

Similar swap properties can be given for other models in the co-occurrence format, e.g., Glove~\cite{glove}.

\begin{observation}
\label{obs:obs2}
Note that Observation~\ref{obs:swap} does not mean that $\W=\C$. If $\ell$ had been a convex function, then, using the fact that any convex combination of optimal solutions is also optimal, one can show the existence of an optimal solution with $\W=\C$. However, the objective function $\ell$ in (\ref{ellall}) is far from convex. Such non-convex objectives are usually associated with optima in which the symmetric components take very different values. In general, even when the symmetry condition does not hold, this discussion indicates that SGNS can end up at different optima depending on the initialization of the optimization process.

\end{observation}

\begin{observation}
\label{predict}
Mikolov et al.~\cite{Mikolov2013b} derive~(\ref{ellall}) starting from the problem of predicting a context word given a word. However, when using negative sampling, the symmetry condition implies that we get exactly the same formulation and solution as for a model where we would predict a word given a context word. It is useful to note that this comment does not hold if traditional softmax or hierarchical softmax~\cite{Mnih2009} or noise-contrastive estimation~\cite{Mnih2013} is used to model probabilities with the associated negative likelihood loss.
\end{observation}

\section{Connecting count and predict models}
In this section, we revisit the idea that Levy and Goldberg~\cite{Levy2014} used to connect PMI models with SGNS and extend it. The central step is to explicitly solve SGNS in closed form; we can then see that the explicit solution of SGNS corresponds to the vectors that are constructed in a Shifted PMI model.

\subsection{Solving SGNS in closed form}
\label{quadratic}
Here, we extend the analysis of Levy and Goldberg~\cite{Levy2014}. We provide closed-form solutions for SGNS and a broad class of loss functions. As Levy and Goldberg showed, it turns out that the closed-form solution of the SGNS objective is equivalent to a solution constructed by the Shifted PMI model.
In contrast to Levy and Goldberg, we apply approximate quadratic analysis to the SGNS objective to give a more detailed insight for a broad class of loss functions.

The central idea of the analysis is to assume that, given a sufficient number of dimensions, each score $x=\w\cdot \c$ can be minimized independently of all other scores. Let us define 
\begin{equation}
\rho_{w,c}(x) = \ell_{w,c}(\w,\c)=\rho_{w,c}(\w\cdot\c).
\label{rhwc}
\end{equation}
To get $x_{w,c}^*$, the minimizer of $\rho_{w,c}$, we solve
\begin{equation}
\rho_{w,c}^\prime(x) = \mp(w,c) L'(x,1) + k\cdot \frac{\mp(w,\cdot)\cdot\mp(\cdot,c)}{|D|} L'(x,-1) = 0
\label{rhwcpr}
\end{equation}
The Taylor series around $x_{w,c}^*$ is given by
\begin{align}
\rho_{w,c}(x) &= \mbox{const.} + \frac{1}{2} \alpha_{w,c} (x - x_{w,c}^*)^2 \label{rtsg} \\
\alpha_{w,c} &= (\mp(w,c) L''(x_{w,c}^*,1) + \frac{k\cdot\mp(w,\cdot)\cdot\mp(\cdot,c)}{|D|} L''(x_{w,c}^*,-1)) \label{alphag}
\end{align}
which, in terms of $\ell_{w,c}$, is
\begin{eqnarray}
\ell_{w,c}(\w,\c) = \mbox{const.} + \frac{1}{2} \alpha_{w,c} (\w\cdot \c - x_{w,c}^*)^2 \label{tsg}
\end{eqnarray}
Let
\begin{equation}
\label{delta}
\delta_{w,c} = \mp(w,c) + \frac{k\cdot\mp(w,\cdot)\cdot\mp(\cdot,c)}{|D|}
\end{equation}
For the loss functions listed in Table~\ref{tab:losslist}, $\alpha_{w,c}$ takes the form
\begin{equation}
\label{alphag1}
\alpha_{w,c} = \gamma_{w,c}\delta_{w,c}
\end{equation}
Table~\ref{tab:losses} gives details for the various loss functions.

\begin{observation}
\label{obs:blowup}
When $\mp(w,c)=0$, the first terms in (\ref{ellwc}), (\ref{rhwcpr}) and (\ref{alphag}) involving $L(x,1)$ go away, causing $x_{w,c}^*$ to take the extreme value of $-\infty$ for the logistic loss and $-1$ for other losses. Consider the expressions for $x_{w,c}^*$ in Table~\ref{tab:losses} to verify this. This issue does not arise for SGNS because it operates with reduced dimensional embeddings of words and context words in which information associated with various (word, context word) pairs interact heavily. We believe that this is a crucial difference between SGNS and PMI methods.
\end{observation}

\begin{table}[h]
\caption{Expressions for $x_{w,c}^*$ and $\alpha_{w,c}$ for various losses. {\em PosCondition} refers to the condition under which $x_{w,c}^* \ge 0$.} 
\label{tab:losses}
\begin{center}
\begin{tabular}{|c|c|c|c|}
\hline
L &  $x_{w,c}^*$ & $\alpha_{w,c}$ & {\em PosCondition} \\
\hline
\hline
Logistic & $\pmi - \log k$ & $\sigma(x_{w,c}^*) \sigma(x_{w,c}^*) \delta_{w,c} $ & $\Pmi > \log k$ \\
\hline
Squared & $\frac{e^{\pmi}-k}{e^{\pmi}+k}$ & $\delta_{w,c} $ & $\Pmi > \log k$ \\
\hline
Squared Hinge & $\frac{e^{\pmi}-k}{e^{\pmi}+k}$ & $\delta_{w,c} $ & $\Pmi > \log k$ \\
\hline
Hinge & $1$ if $\pmi\ge \log k$ & NA & $\Pmi > \log k$ \\
      & $-1$ otherwise         &    & \\
\hline
Huber & $\frac{e^{\pmi}-k}{e^{\pmi}+k}$ & $\delta_{w,c} $ & $\Pmi > \log k$ \\
\hline
\end{tabular}
\end{center}
\end{table}

\begin{observation}
\label{obs:xalosses}
For building semantic representations that are used for computing similarity, it is often not desirable to have negative components in the word vectors. PosCondition in the last column of Table~\ref{tab:losses} checks when this case occurs. This condition turns out to be the same for all losses, which is interesting. There is no difference between squared loss, squared hinge loss and Huber loss. This is because, in the interval $x\in [-1,1]$, all these losses have identical $\rho_{w,c}(\cdot)$ and the minimizer of $\rho_{w,c}$ always occurs in $[-1,1]$. These three losses have an expression for $x_{w,c}^*$ that is quite different from that for the logistic loss. Hinge loss prefers to set $x_{w,c}^*$ at the extremes: $1$ or $-1$. 
\end{observation}

\subsection{Connecting SGNS and Shifted PMI}
\label{subsec:connectsgns}
One of the key results of Levy and Goldberg~\cite{Levy2014} is that the vectors created by Shifted PMI are a solution to the SGNS objective. We use the quadratic analysis of Section~\ref{quadratic} to say this more cleanly. 


Let $\e_i$ denote the unit vector in $\mathbb{R}^{|\V|}$ whose $i$-th component is $1$ and all other components are zero. Suppose we use a one-hot representation for context vectors, i.e., we fix $\c = \e_c$ for all $c$. Thus, we are fixing $\C$.

\begin{observation}
\label{ob:minpmi}
Suppose we fix each context vector to the one-hot representation given above. Then, only $\W$ remains as the set of variables, and the following hold.
\begin{enumerate}
\item[(a)] The minimizer $\W^*$ of (\ref{ellwc}) is given by $\w = \x_w^*$ $\forall w$, where $\x_w^*$ is a vector with $\{x_{w,c}^*\}_c$ as components.
\item[(b)] $x_{w,c}^*$ is the Shifted PMI representation as defined in Section~\ref{sec:pmi}. 
\item[(b)] Also, $(\W^*,\C)$ together form an optimal solution of SGNS.
\end{enumerate}

\end{observation}

In other words, we have a closed-form solution for SGNS (assuming $m = n = |\V|$). Though Levy and Goldberg~\cite{Levy2014} do not mention the above construction, this is a simple observation that easily follows from their analysis.

{\bf Proof of Observation~\ref{ob:minpmi}.}
Let's take one $c$. By the way we defined $\c$, we have $\w\cdot\c = \w_c$. In $\ell$ given by (\ref{ellall}), the variable $\w_c$ occurs only in the term $\ell_{w,c}$. Therefore, $\w_c^* = \arg\min \ell_{w,c} = x_{w,c}^*$, which proves part (a) of Observation~\ref{ob:minpmi}. Part (c) follows from the fact that, the pair, $(\W^*,\C)$ is such that $\ell_{w,c}$ is minimized for every $(w,c)$ pair and it is not possible to do better than that.

\section{Count and predict models: differences}
\label{sec:connect}



Empirically, there seems to be evidence that predict and count models perform differently (e.g., \cite{Baroni2014}). This is interesting since they all consider the same input data -- namely the co-occurrences of words in text. What are the reasons for this? 
Although the previous section pointed out a strong connection between SGNS and PMI methods (with Observation~\ref{ob:minpmi} even indicating a near equivalence), we believe there are two main differences between the two methods pertaining (a) the dimension of the embeddings, and (b) $\C$ being fixed as the one-hot representation. Let us now expand on the two factors.

\begin{observation}
\label{obs:dim}
{\bf Dimension of embeddings.}
As already mentioned in Observation~\ref{obs:blowup}, the small embedding dimension used by SGNS for words can be crucial for learning good word vectors. For example, co-occurring words can influence each other. The full dimension used by PMI methods does not allow this to happen; the full dimension also has the disadvantage of suffering from overfitting due to an excessive number of variables.
\end{observation}

\begin{observation}
\label{obs:onehot}
{\bf One-hot representation for $\C$.}
Similar to the previous observation, there is a difference in which variables are optimized. 
SGNS operates with both, $\W$ and $\C$ as variables. As shown in Subsection~\ref{subsec:connectsgns}, PMI methods, on the other hand, correspond to using a fixed one-hot representation for $\C$. An important consequence of such a representation is that it does not allow close context words to influence each other. Future work should empirically investigate the role of this factor.

\end{observation}



\subsection{Further differences in Shifted PPMI}
Recall that the solution for SGNS given by Shifted PMI is unusable in practice, since we have entries that are $-\infty$. Also, the Shifted PMI solution has a high number of dimensions, and this might not be useful in practice. Levy and Goldberg present two heuristics remedy these problems. First, they propose omitting negative terms in the objective function to make a solution feasible. Second, they suggest a subsequent SVD step to reduce the dimensionality of the Shifted PPMI matrix. We below discuss each of these heuristics and their consequences in more detail. 

\subsubsection{Omitting terms}
\label{obs:discard}
Since in practice, we cannot work with vectors that have $-\infty$ entries, Goldberg and Levy propose Shifted PPMI instead of Shifted PMI to remedy this problem. Shifted PPMI corresponds to leaving out all $(w,c)$ terms from (\ref{ellall}) that have negative Shifted PMI values. This is equivalent to a modified SGNS method in which during negative sampling examples not satisfying the PosCondition, i.e., those with $\pmi - \log k<0$ are left out.

There is two issues with the above approach. First, we no longer can guarantee optimality for this solution. Second, this also seems inconsistent with the main idea behind negative sampling, which is to sample from unobserved $(w,c)$ pairs. Levy and Goldberg~\cite{Levy2014} make the following statement in the second paragraph of Section 5.1 of their paper: ``{We observe that Shifted PPMI is indeed a near-perfect approximation of the optimal solution, even though it discards a lot of information when considering only positive cells.}" However, this does not explain the role of the discarded terms. In particular, when training SGNS with a low number of embedding dimensions, discarding those terms could be of real importance.

\subsubsection{Applying SVD}
Levy and Goldberg also propose to apply SVD to the SPPMI matrix in order to obtain low-dimensional embeddings. If we follow this step, we loose again some of the optimality we had with the Shifted PMI solution.
More formally, consider a SPPMI matrix $M$ whose $(w,c)$'th term is $\max(0, x_{w,c}^*)$. To form word vectors of dimension $d$ lower than $|\V|$, Goldberg and Levy suggest to apply SVD to the matrix $M$.  Let $M = U\Sigma V^T$ denote the SVD. If one is interested in an embedding of dimension $d<|\V|$, the best rank $d$ approximation of $M$, $U_d\Sigma_d V_d^T$ is used. To form word vectors, one can use either $\W_d^\svd =U_d\Sigma_d$ or the ``symmetric version", $\W_d^\ssvd =U_d \sqrt{\Sigma_d}$. 

\begin{observation}
\label{obs:svd}
Levy and Goldberg~\cite{Levy2014} propose that SVD is done on $M$ or one of its variants.\footnote{Levy and Goldberg~\cite{Levy2014} recommend using the matrix corresponding to Shifted PPMI.} However, if remaining faithful to SGNS is the aim, (\ref{tsg}) indicates that the weighting term $\alpha_{w,c}$ also be included and that we solve
\begin{eqnarray}
\min_{\W, \C} \frac{1}{2} \sum_{w,c} \alpha_{w,c} (\w\cdot \c - x_{w,c}^*)^2 \label{wsvd}
\end{eqnarray}
It is non-trivial to solve this problem; an SVD based approach will not work. 
\end{observation}

Let us now look at the limiting full case, i.e., $d=|\V|$, to point out some relations and differences within the methods in the PMI class.

\begin{observation}
\label{obs:fullsvd}
We can use the full SVD of $M$ and define word vectors $\W_{\full}=U\Sigma$ and $\W_{\full}^{\symm}=U\sqrt{\Sigma}$. Note that these word vectors are not the same as Shifted PPMI which uses $M$ itself as word vectors. However, because $MM^T=U\Sigma V^TV\Sigma U^T=U\Sigma\Sigma U^T=\W_{\full}\W_{\full}^T$, the dot products between any two word vectors is identical for $\W^{\sppmi}$ (i.e., $M$) and $\W_{\full}$. On the other hand, in general, $\W_{\full}^{\symm} (\W_{\full}^{\symm})^T \not= MM^T$.  What this means is that SVD is consistent with Shifted PPMI, but Symmetric SVD is not consistent with Shifted PPMI. 
\end{observation}


Levy and Goldberg~\cite{Levy2014} recommend Shifted PPMI and refer to the spectral word vectors for it as {\em SVD} and {\em Symmetric SVD}. Their experiments showed the symmetric version to yield better results than SGNS.


\subsection{Summary}
\label{subsec:summary}
\begin{figure}[H]
\centering
\includegraphics[width=1\linewidth]{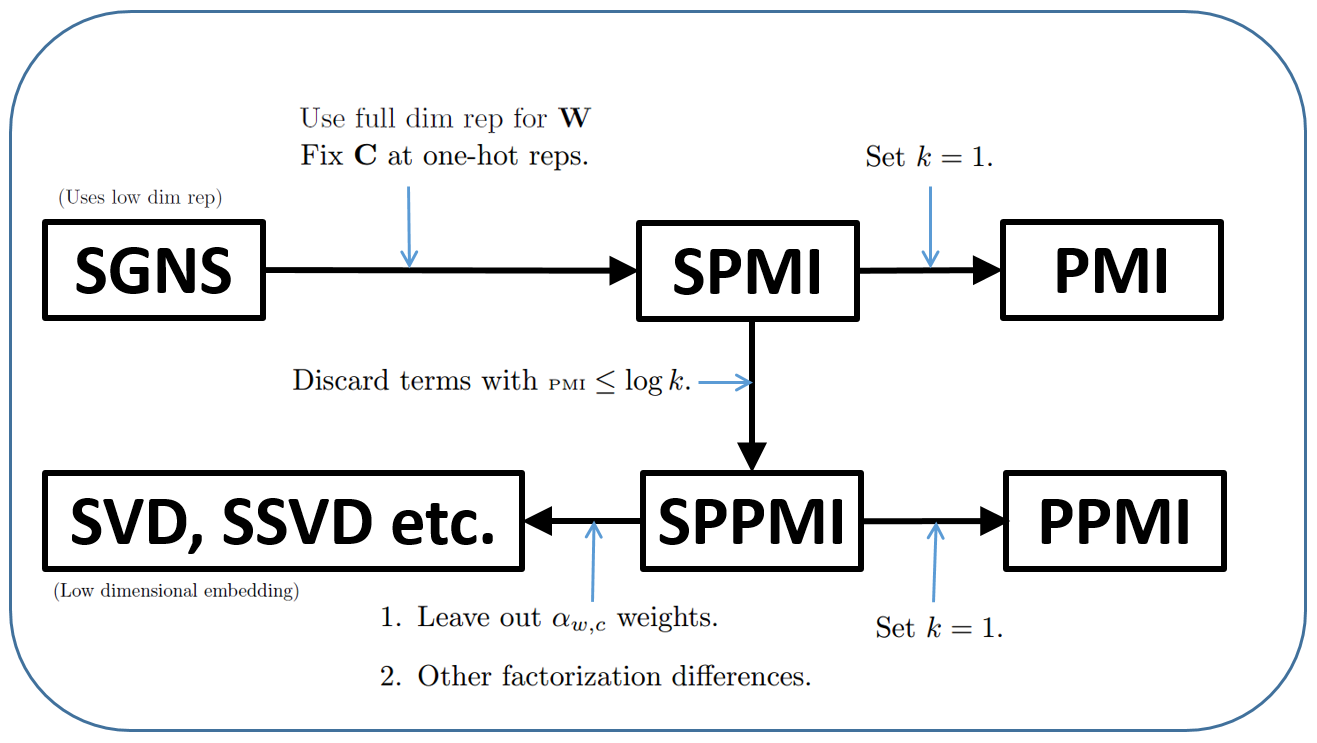}
\caption{Differences between various methods. Here ``discarding" refers to the leaving out of certain negative examples satisfying some condition, after their creation.}
\label{fig:diffs}
\end{figure}
The above subsections went into various ways of connecting SGNS and PMI methods and also brought out various differences. Figure~\ref{fig:diffs} gives a rough and quick view of the various morphings of SGNS into different PMI methods. 
The use of a small embedding dimension by SGNS as opposed to the full dimensional one hot representation used by PMI methods is probably the most important difference.
The discarding of certain negative examples, the approximation of the original non-linear objective by a quadratic, and the leaving out of the $\alpha_{w,c}$ factors from the quadratic objective, are additional differences. Carefully designed experiments are needed to understand the individual effects of each difference on the quality of the resulting word vectors.
The various differences also imply that, even though the methods at the two left ends of the figure involve low dimensional embeddings, they could be vastly different due to the various reasons given in this section.




\section{Extensions}
In the following two subsections, we propose two extensions to the basic PMI model. First, we show how to add regularization to PMI models and give explicit solutions for $L_2$ and $L_1$ regularization. We hope that regularization will help improve issues with data sparsity. After that, we propose a new convex model for word embeddings that is not only easy to learn, but also yields intuitive word vectors since each dimension corresponds exactly to one context word.

\subsection{Adding regularization to PMI methods}
\label{sec:addreg}

Let us continue with the formulation of Section~\ref{quadratic} and add regularization. If overfitting is one of the causes of the inferior performance of PMI methods compared to SGNS, then regularization should help improve performance. Consider the decoupled determination of $x=\w_c$. Our modified objective is now $$\argmin_x \rho_{w,c}(x) + R_{w,c}(x)$$ where $R_{w,c}$ is the regularization term. As we argued in Section~\ref{quadratic}, each $\w_c$ is decoupled from other variables. Hence, we can solve each one-dimensional optimization problem in isolation -- even with regularization added. We now derive closed-form solutions for $L_2$ and $L_1$ regularization.

\subsubsection{$\bf L_2$ regularization}
\label{subsec:l2regsoln}
\begin{figure}[H]
\centering
\includegraphics[width=1\linewidth]{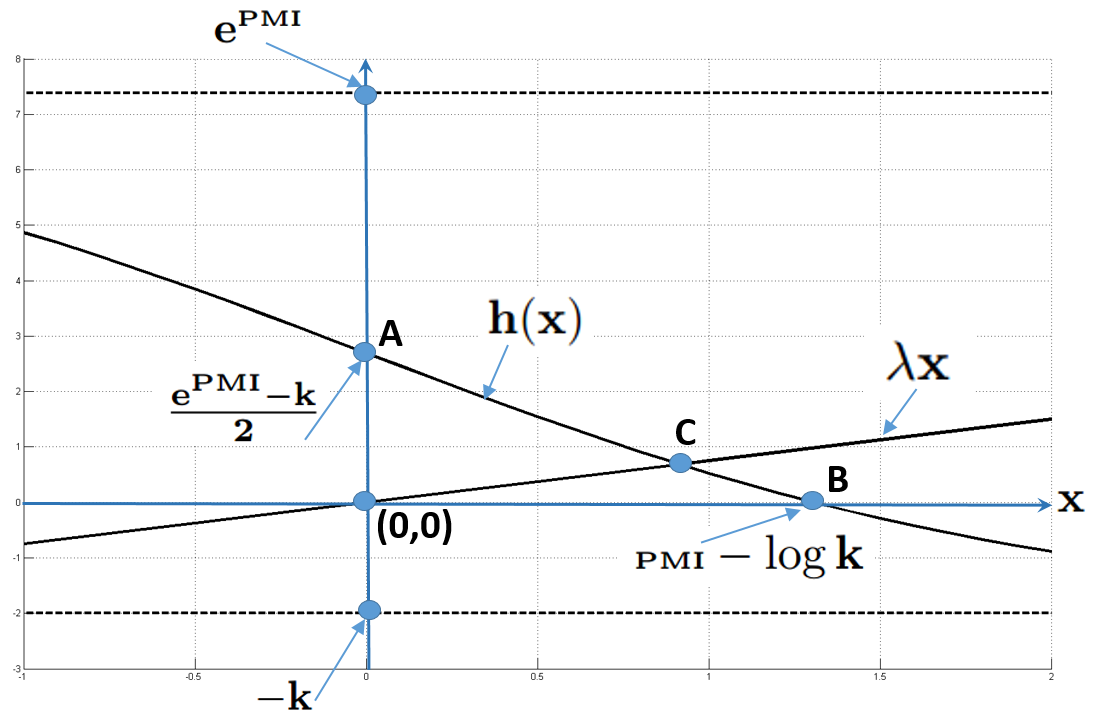}
\caption{The determination of the $L_2$ regularized solution. Note that, even though $h(x)$ is a nonlinear function, the straight line from A to B approximates that part of the curve well.}
\label{fig:l2reg}
\end{figure}

We can down-weigh frequently occurring words and context words and add a standard $L_2$ regularizer as follows:
\begin{equation}
\label{rwc}
R_{w,c} (x) = \lambda \frac{\mp(w,\cdot)\cdot\mp(\cdot,c)}{|D|} R(x) \mbox{   where   } R(x) = \frac{1}{2} x^2
\end{equation}
We will discuss $L_1$ regularizer $R(x)=|x|$ later. Let us also focus only on the logistic loss. Dividing by $\frac{\mp(w,\cdot)\cdot\mp(\cdot,c)}{|D|}$ we get
\begin{equation}
\label{wcstar}
\w_c^* = \arg \min_x e^{\pmi} \log (1+e^{-x}) + k \log (1+e^x) + \frac{\lambda}{2} x^2
\end{equation}
Setting the derivative of the objective to zero and simplifying, we get $\w_c^*$ to be the solution of
\begin{equation}
\label{hx}
\lambda x = h(x) \mbox{   where   } h(x) = \frac{e^{\pmi} - k e^x}{1+e^x}.
\end{equation}
The line from $A$ to $B$ in Figure~\ref{fig:l2reg} approximates $h(x)$ nicely in the region where the optimal $x$ lies. The equation of this line is given by 
\begin{equation}
\label{htilde}
\tilde{h}(x) = \frac{e^{\pmi}-k}{2} - \frac{e^{\pmi}-k}{2(\pmi-\log k)} x
\end{equation}
Solve for $\tilde{h}(x)=\lambda x$ to get the optimal $x$ as
\begin{equation}
\label{hmean}
x = \mbox{ Harmonic mean of } (\pmi-\log k, \frac{e^{\pmi}-k}{2\lambda})
\end{equation}

\subsubsection{$\bf L_1$ regularization}
\label{subsec:l1regsoln}

Similarly, it is easy to work out closed form expressions for $\w_c^*$ for $L_1$ regularization. In this case, many of the optimal values go to zero: the larger the value of $\lambda$, more is the number of zeros.
There are three cases to consider. For $L_1$ regularization, the value, $h(0)=\frac{e^\pmi-k}{2}$ plays a key role.

{\bf Case 1.} ${\bf -\lambda \le h(0) \le \lambda}$. For this case it is easy to check that, at $x=0$, the left hand side derivative of the objective in (\ref{wcstar}) is non-positive, and the right hand side derivative is non-negative. Hence $\w_c^*=0$ is the optimal solution.

{\bf Case 2.} ${\bf h(0) > \lambda}$. For this case, the right hand side derivative at $x=0$ is negative and so the optimum value is positive. The optimum is found by solving $h(x)=\lambda$, which yields 
\begin{equation}
\w_c^* = \log \frac{e^\pmi - \lambda}{k+\lambda}
\label{case2}
\end{equation}

{\bf Case 3.} ${\bf h(0) < -\lambda}$. For this case, the left hand side derivative at $x=0$ is positive and so the optimum value is negative. The optimum is found by solving $h(x)=-\lambda$, which yields 
\begin{equation}
\w_c^* = \log \frac{e^\pmi + \lambda}{k-\lambda}
\label{case3}
\end{equation}

\subsubsection{Discussion}
\label{subsec:l1l2comments}

Adding regularization to count models has two benefits. First, while the non-regularized solution (see Table~\ref{tab:losses}; same as SPMI) is unusable because $x_{w,c}^*$ goes to extreme values when $\mp(w,c)=0$, regularized solutions are always well-defined. Second, we expect regularization to help if overfitting is degrading performance with PMI models. Future work is needed to empirically study and validate the usefulness of regularization in count models. 

\subsection{A convex formulation for word vectors}
\label{sec:convex}

Motivated by the analysis in Section~\ref{quadratic}, we develop a new and simple convex formulation for determining word vectors. Similar to the skip-gram model, we phrase our model as the task of predicting a word conditioned on context. However, instead of learning representations for both words and their context words, we fix the context vectors to a one-hot representation. This makes our objective function \emph{convex}. Another advantage is that this results in \emph{intelligible} models -- each vector entry refers to a weight given to a context word. Also, to obtain compact representations, we instead use $L_1$ regularization on the word vectors instead of learning embeddings in lower-dimensional spaces like traditional predict models.
 
Let us describe the formulation in some more detail. Let $m$ denote the dimension of the vector representation of context. Depending on the context modeling employed, $m$ can be one to several times of $|\V|$. Let $\z\in \mathbb{R}^m$ denote the context representation. $\z$ is a function of the context, $C$, written as $\z(C)$. The weight vectors $\{\w\}$ are also points in $\mathbb{R}^m$. The score of a positive example is $\w\cdot\z(C)$. Remember that the input corpus $D$ is a sequence of words, $\{w^t\}_{t=1}^T$. Let the corresponding sequence of contexts be $\{C^t\}_{t=1}^T$. We solve the following minimization problem.
\begin{equation}
\min_{\W} \sum_{w=1}^{|\V|} \lambda \|\w\|_1 + \frac{1}{T} \sum_{t=1}^T f(\w^t;\z(C^t))
\end{equation}
In the above, $\lambda>0$ is a regularization constant\footnote{One can also think of schemes for making $\lambda$ dependent on $w$, for example, make $\lambda_w$ dependent on the number of context words to which word $w$ is attached.} which can be tuned to balance sparsity of $\W$ and performance; $f$ includes the effect of positive as well as negative examples. Other forms of regularization are also possible, for example a combination of $L_1$ and $L_2$ regularization (Generalized Elastic net).

{\bf Context.} We can define context in several ways. Here are a few possibilities:
\begin{enumerate}
\item A simple way is to have a context $\z\in\mathbb{R}^{|\V|}$, with each dimension corresponding to one word in the vocabulary $V$. Similar to the SGNS model, we only assume \emph{individual} interactions between words and context pairs, i.e., each $\z$ will exactly contain one non-zero entry. This way, we can rewrite the objective in the co-occurrence form and obtain all vectors in closed form by accumulating the statistics for each (word, context word) pair. 
\item We can do better than option 1 above and pair a word with a window of context words {\em simultaneously}, i.e., $\z$ can now have multiple non-zero entries. A bag-of-words representation $\z\in R^{|\V|}$ can be used to represent the context input using the context words. For example, we can set $\z_c=\rho(c,w)$ if $c$ occurs in the context window and zero otherwise, where $\rho(c,w)$ is some weighting that is a function of how far $c$ is from $w$. 
\item If we want to make use of word order, we can define $L\times |\V|$ context inputs where $L$ is the length of the context window and one block of $|\V|$ inputs is used for each context word, and blocks concatenated in the proper order of occurrence in the window.
\item Depending on the purpose for which the word vectors are developed we can also use dependency-tree based context features~\cite{Levy2014a}. Again, we can use one-hot representations of these features.
\end{enumerate}

{\bf Learning.} Except option 1, aggregating at the individual $(w,c)$ level (to convert the objective into the co-occurrence form~(\ref{ellall})) as well as forming a closed form solution, is in general infeasible. This means we have to train our model using the objective in \emph{corpus format}. To do the training, we can choose from various optimization strategies. We can employ: (a) proper softmax over all words; (b) hierarchical softmax~\cite{Mnih2009}; (c) Noise-contrastive estimation (NCE)~\cite{Mnih2013}; or, (d) negative sampling like in SGNS~\cite{Mikolov2013a}.

We now present examples for two of the four optimization strategies mentioned above. With softmax, our objective is
\begin{equation}
f(\w^t;\z(C^t)) = -\w^t\cdot\z(C^t) + \log \left(\sum_w \w\cdot\z(C^t)\right).
\label{softmax}
\end{equation}
Let us now consider optimization via Negative Sampling. Let $\{w_{\neg}^j\}_{j=1}^k$ denote a randomly sampled set of negative examples. Then 
\begin{equation}
f(\w^t;\z(C^t)) = -\log \sigma (\w^t\cdot\z(C^t)) - \sum_{j=1}^k \log \sigma (-\w_{\neg}^j\cdot\z(C^t)),
\label{negsamp}
\end{equation}
where $\sigma(x) = 1/(1+e^{-x})$.

\begin{observation}
\label{obs:interp}
One of the advantages of the convex formulation described above is {interpretability}. If the $r^{\text{th}}$ component of some $\w$ is large and positive, it can be directly translated to the role played by the $r^{\text{th}}$ context term of $\z(C)$. This direct interpretability also means that domain knowledge may be easy to infuse into the formulation by specification of constraints and extra regularization on the weights.
\end{observation}

\begin{observation}
\label{highorder}
The model developed in this section is in between conventional low-dimensional word embeddings (no interpretability, allows for higher-order effects) and distributional word representations (full interpretability, no higher-order effects). Here, the term ``higher-order effects" refers to the effects (influences) that occur during the joint optimization of context and word representations, e.g, two words can be made similar because they occur in contexts that are also similar.
\end{observation}

\section{Conclusion}
\label{sec:conc}
In this report, we showed how to explicitly solve the SGNS objective for a broad range of loss functions. This step allowed us to connect Shifted PMI models to SGNS models under general loss functions. Furthermore, we pointed out two important differences between the Shifted PMI model and SGNS model. First, in the SGNS model, far fewer embeddings are used in practice, making the SGNS model less prone to overfitting. Second, in the PMI model, the context vectors are fixed; hence, there is also no interaction between context vectors and word vectors. 

Finally, we proposed two extensions to existing models. First, we showed how we can incorporate regularization into PMI models to alleviate overfitting. Second, we presented a new embedding model that not only has a convex objective, but also results in intelligible embeddings. Future work is needed to empirically validate our proposed methods and extensions.


\begin{thebibliography}{1}

\bibitem{Baroni2014} M. Baroni, G. Dinu and G. Kruszewski. {\em Don’t count, predict! A systematic comparison of context-counting vs. context-predicting semantic vectors.} ACL, 2014.

\bibitem{Levy2014} O.~Levy and Y.~Goldberg. {\em Neural word embedding as implicit factorization.} NIPS 2014.

\bibitem{Levy2014a} O.~Levy and Y.~Goldberg. {\em Linguistic Regularities in Sparse and Explicit Word Representations.} CoNLL, 2014.

\bibitem{Levy2015} O.~Levy, Y.~Goldberg and I.~Dagan. {\em Improving distributional similarity with lessons learned from word embeddings.} TACL, 2015.

\bibitem{Mikolov2013a} T. Mikolov, K. Chen, G. Corrado, and J. Dean. {\em Efficient estimation of word representations in vector space.} CoRR, abs/1301.3781,
2013.

\bibitem{Mikolov2013b}T. Mikolov, I. Sutskever, K. Chen, G.S. Corrado, and J. Dean. {\em Distributed representations of words and phrases and their compositionality.} NIPS, 2013.

\bibitem{Mnih2009} A. Mnih and G. Hinton. {\em A scalable hierarchical distributed language model.} NIPS, 2009.

\bibitem{Mnih2013} A. Mnih and K. Kavukcuoglu. {\em Learning word embeddings efficiently with noise-contrastive estimation.} NIPS, 2013.

\bibitem{glove} J. Pennington, R. Socher,   C. D. Manning. {\em GloVe: Global vectors for word representation.}  EMNLP 2014.

\bibitem{Shi2014} T.~Shi and Z.~Liu. {\em Linking Glove with} {\tt word2vec}.arXiv: 1411.5595v2, 2014.

\bibitem{Stratos2015} K.~Stratos, M. Collins and D.~Hsu. {\em Model-based Word Embeddings from Decompositions of Count Matrices.} ACL, 2015.

\bibitem{Turney2010} P.~Turney and P.~Pantel. {\em From frequency to meaning: Vector space models of semantics.} JAIR, 2010.


\end{thebibliography}
\end{document}